\documentclass[sigconf,nonacm]{acmart}

\usepackage{natbib}
\setcitestyle{square, comma, numbers,sort&compress, super}

\begin{document}

\title{CPST: Comprehension-Preserving Style Transfer for Multi-Modal Narratives}


\author{Yi-Chun Chen}
\affiliation{%
  \institution{Computer Science, North Carolina State University}
  \country{USA}
  }
\email{ychen74@ncsu.edu}

\author{Arnav Jhala}
\affiliation{%
  \institution{Computer Science, North Carolina State University}
  \country{USA}
  }
\email{ahjhala@ncsu.edu}

\renewcommand{\shortauthors}{Chen et al.}

\begin{abstract}
We investigate the challenges of style transfer in multi-modal visual narratives. Among static visual narratives such as comics and manga, there are distinct visual styles in terms of presentation. They include style features across multiple dimensions such as panel layout, size, shape, and color. They include both visual and text media elements. The layout of both text and media elements is also significant in terms of narrative communication. The sequential transitions between panels is where readers make inferences about the narrative world. These feature differences provide an interesting challenge for style transfer in which there are distinctions between the processing of features for each modality. We introduce the notion of comprehension-preserving style transfer (CPST) in such multi-modal domains. CPST requires not only traditional metrics of style transfer but also metrics of narrative comprehension. To spur further research in this area, we present an annotated dataset of comics and manga, and an initial set of algorithms that utilize separate style transfer modules for the visual, textual, and layout parameters. To test whether the style transfer preserves narrative semantics, we evaluate this algorithm through visual story cloze tests inspired by work in computational cognition of narrative systems. Understanding the connection between style and narrative semantics provides insight for applications ranging from informational brochure designs to data storytelling.
\end{abstract}

\maketitle

\section{Introduction}


The neural style transfer problem has been an active topic of research with notable results in static images \cite{gatys2015neural}. Results on transfer of artistic elements such as color and stroke textures has been impressive. There has been steady progress both in terms of quality and efficiency of these algorithms. These results are useful for appealing to human viewers at the perceptual level. In this paper, we apply style transfer to sequential narrative medium of comics and manga which has as yet been not systematically explored. This domain poses interesting challenges for style transfer due to the presence of features such as panel size and layout, text and its placement within bubbles, and the visual scene. We show that applying style transfer directly to a comic panel with existing approaches leads to mixing of styles from these three channels. This form also has sequential narrative requirements of maintaining coherence across panels. To address this issue, we introduce the notion of comprehension preserving style transfer between different art-styles of sequential images with narrative content. We provide an initial set up of a dataset with annotations that includes western comics and Japanese manga. We present a computational framework for transferring style across three separate channels of layout, scene, and text. We show visual results of transfer on a set of images and propose a new evaluation method based on visual story-cloze test~\cite{iyyer2017amazing} comparison between the original and style-transferred image sequences. This comprehension test indicates whether coherence between panels is preserved by the style transfer process.  



Comics and manga have distinct art styles that vary across several aspects~\cite{mccloud1998understanding, mccloud2006making}; using simplified depictions to capture complex ideas like object, shape, or shadow, abstract symbols to represent meanings, and expressing temporal and spatial changes through connecting content into a sequence. They use a range of styles from simplified line sketches to realism and share characteristics such as composition, contrast, lighting, and texture with traditional visual art. Due to the narrative content being communicated they are carefully designed to preserve narrative semantics~\cite{pratt2009narrative}. This makes aspects like the relations of entities shared between panels and panel layouts become part of their art style.



\section{Related work}
Visual style transfer is studied as a generalized problem of texture synthesis, which is to extract and transfer the texture from a source to a target image \cite{ ashikhmin2003fast,  shih2014style, zhang2013style, rosin2012image}. It has been used for non-photorealistic rendering \cite{ bruckner2007style, ma2014analogy} to transfer scenes or objects into a particular artistic style. Gatys’ et al.~\citeyear{gatys2016image} first introduced a method for using pre-trained convolutional neural network (CNN) to reproduce famous painting styles on natural images. Neural style transfer since that work has been a widely discussed topic \cite{ jing2019neural}. This original approach has two limitations. First, the training process is not efficient. To address this, feedforward style transfer and other methods then were proposed to accelerate the process \cite{chen2016fast, johnson2016perceptual}. These faster approaches allow multiple images to be processed faster making them feasible for processing multiple sequences of images such as in our dataset. Second, transferring the entire image ignores the relationships between content of the images. This poses a challenge for our application due to the presence of overlaid text within bubbles and the different sizes and layout of panels on a page. One way to approach the content problem is to use masking \cite{shen2016automatic, castillo2017zorn} or to control the areas influenced during the transfer process \cite{gatys2017controlling}. Considering the nature and complexity of our target data we utilize masked variants of images according to content.

In terms of work related to comics, we incorporate two datasets in this project and augment them with annotations. The COMICS dataset which includes 1.2 million panels of western comics \cite{iyyer2017amazing}. The Manga109 dataset includes 109 titles of Japanese manga \cite{narita2017sketch, matsui2017sketch}. Our method is inspired by other content-aware style transfer, comic sense-making \cite{augereau2018survey, martens2020visual}, and content understanding with cloze-tasks and fine-tuned features \cite{park2019estimating} on narrative understanding.

\section{Datasets}

Two datasets are chosen to demonstrate the style transfer approach in this paper. The COMICS dataset developed by Iyyer et al. (\citeyear{iyyer2017amazing}) is chosen as the source dataset for style. This dataset includes 3,948 western comics with diversity in visual style and genres, separated into 1.2 million panels in total. COMICS contains separated panel images plus around 500 raw pages that preserve the original layouts of comic style. The Manga109 dataset includes 109 titles of manga, which is a specialized art style of Japanese comics. The 109 titles were created by professional comic artists and were all published in commercial manga magazines between the 1970s and 2010s that encompass a wide range of genres. The dataset includes entire pages for each manga title. It also includes annotations of textbox position, text content, and positions of character bodies and faces on the page. Reading order in manga is different from comics. Panels are to be read right to left and top to bottom. We labeled and analyzed the reading order of panels to get the correct sequences. Manga also contains panels sizes and placement that are different from comics that are mostly grid-based.


\section{Approach}
\begin{figure*}[ht]
    \centering
    \includegraphics[width=16cm]{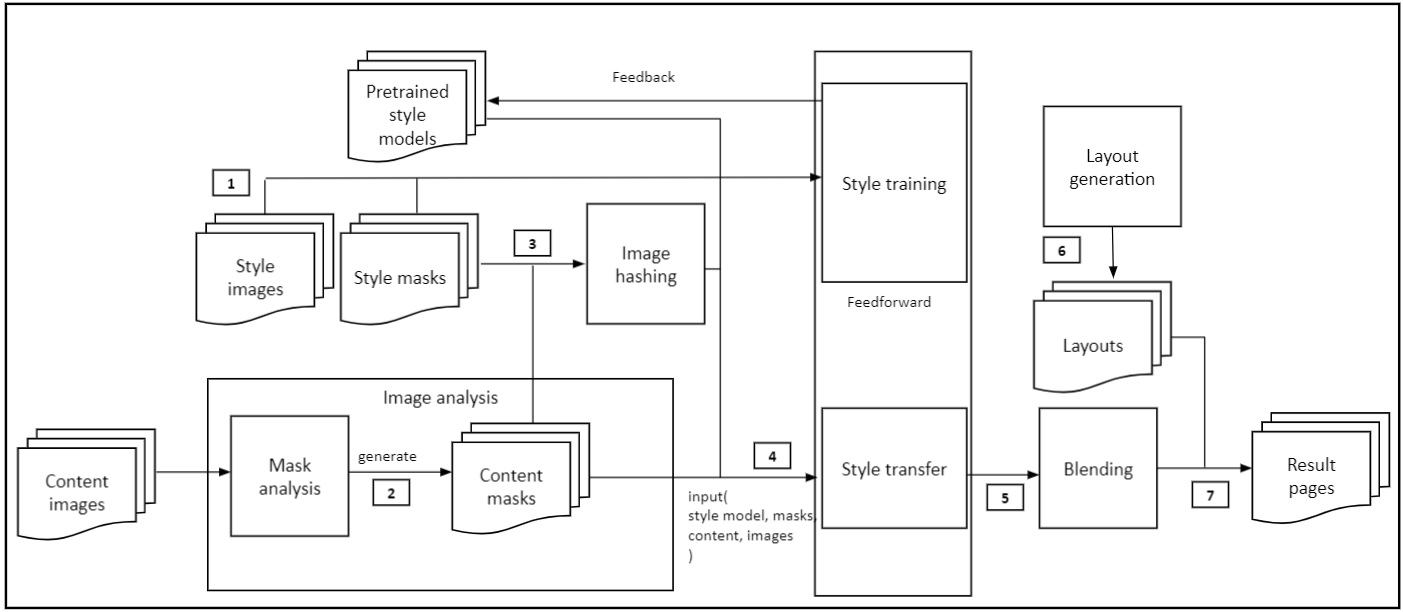}
    \caption{The framework for the comic style transfer process. [1] destination style images and masks are used for training; [2] input images are masked; [3] inputs' masks and destination images' masks are hashed, and style are selected by similarity; [4] style transfer module transfers channels in parallel; [5] per-channel outputs are blended to form a single image;[6] generate comic layout; [7] output images are stored and combined with layout}
    \label{fig:pipeline}
\end{figure*}

The process of style transfer requires processing of features of the target style trained on a dataset of target images and the processing of input image to be converted. The target style in our work is the western comic style and the input images are in manga style. We choose this setup because manga images do not have color making the comics to manga conversion less interesting for demonstration. Comparison of transfer in both directions is outside the scope of this paper and is planned as the next study in this project. Processing input features for target style is challenging because comic pages are complex. The scene, text, and layout layers have their own style features that are perceptually independent. To address this issue, we create masked versions of each panel image with layout feature annotations. These masked images are then passed through separate style transfer pipelines. For the input image to be converted, similar masking is done and the style is transferred to corresponding masked images. The style-converted images from channels are finally blended together to construct the output image.

\subsection{Computational Framework}


Figure~\ref{fig:pipeline} shows the overall framework for style transfer. Our model follows the approach from Johnson et al.(\citeyear{johnson2016perceptual}) and Gordic et al.(\citeyear{Gordic2020nstfast}), where is a feedforward style transfer (FFST).  We try the general CNN style transfer model \cite{gatys2015neural} before we choose this; however, the time cost is extremely high to get converted results for tens-thousands of manga panels. Thus, we use FFST, which reached similar qualitative results but is three orders of magnitude fast.

It has two components: image transformation network and loss network. The image transformation network is a deep residual convolutional neural network parameterized by weights. It maps input images into output images. The loss functions measures the difference between the output image and destination style or content images. They are defined through the loss network. The image transformation network is trained using the Adam algorithm, a replacement optimization algorithm for stochastic gradient descent to train deep learning models when handling sparse gradients on noisy problems. This is useful because masked images cause sparsity on feature representations. The optimizer minimizes a weighted combination of loss functions.

 To better measure perceptual and semantic differences between images, the loss network is a 16-layer convolutional neural network \cite{simonyan2014very} pre-trained on ImageNet (VGG-16) \cite{russakovsky2015imagenet}. This is because it has already learned the perceptual and semantic information we want to measure in the loss functions. ImageNet is trained on real images and it has been documented that it is texture-biased\cite{chen2020shape} while our application is on stylized images. We acknowledge this as an issue. Future work could look at more specialized approaches and incorporate models that better fit this problem to improve benchmark performance.

The loss network defines two perceptual loss functions in the style transfer problem: feature reconstruction loss and style reconstruction loss.  The former is the feature difference between the output image and the content image, and the latter is the difference between the output image and the style image. They both use the feature representations in the different convolutional layers of the loss network to compute the loss. The squared and normalized Euclidean distance between the output image's feature representation and the content image's is the feature reconstruction loss. 

Given a feature map with $C_{j} \times H_{j}, \times W_{j} $ dimensions that represents the input images, where the $H_{j}$ and $W_{j}$ are the height and weight, and each point on its grid is described as $C$-dimensional features.Here is how the style loss is defined. First, a gram matrix is defined as a $C \times C$ matrix whose elements are the product of the feature matrix and then normalized by dividing with its dimension. This denotes the proportional to the uncentered covariance of the  $C$ dimensional features. Style reconstruction loss is defined as the squared Frobenius norm of the difference between the gram matrices of the output image and style image.

\subsubsection{Image composition and masking}
Due to multi-modalities in manga, a panel can be split into two parts: scene content that carries visual information and textboxes that convey textual information. The scene content can be further divided into foreground and background. The major object that performs the story is counted as the foreground and the scene that carries additional information is the background. Based on this, we create three different masked images from the source image. Images with text bubbles masked, images with foreground characters masked, and images with background masked. These masked images are used to train independent style models. Examples are in table \ref{masks}.

\begin{table}
\centering
\begin{tabular}{|p{1.8cm}|p{1.8cm}|p{1.8cm}|p{1.8cm}|}
    \hline
    Image & Textbox mask & Foreground mask & Background \\
    \hline
    \includegraphics[width = \linewidth]{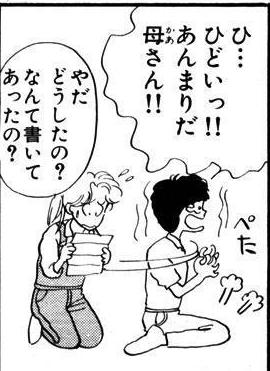}
    &\includegraphics[width = \linewidth]{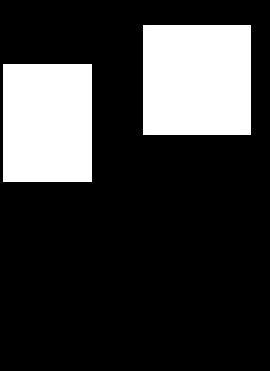}
    &\includegraphics[width = \linewidth]{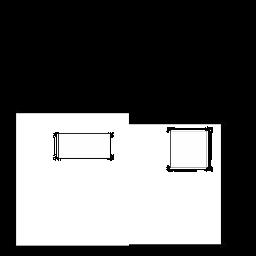}
    &\includegraphics[width = \linewidth]{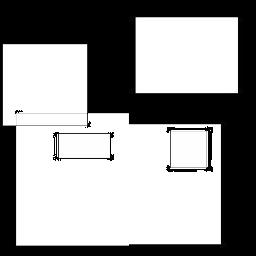}\\
    \hline 
    \includegraphics[width = \linewidth]{Images/5_5_panel.jpg}
    &\includegraphics[width = \linewidth]{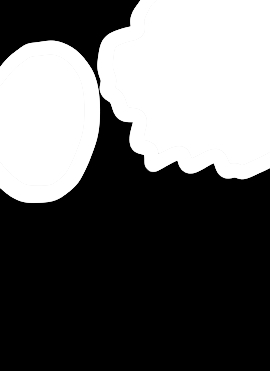}
    &\includegraphics[width = \linewidth]{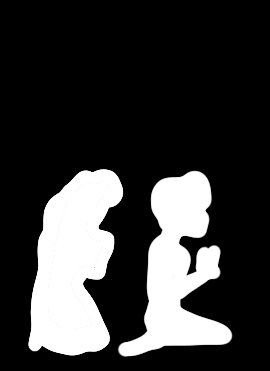}
    &\includegraphics[width = \linewidth]{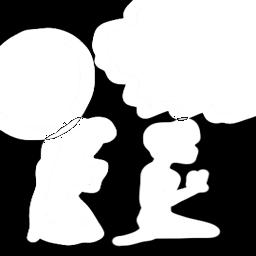}\\
    \hline     
\end{tabular}
    \caption{
    \label{masks} \small{Examples of both rectangle and fit masks for textbox, foreground, and background masks are their combination.}
    }
\end{table}

After comparing the similarity between masks of both content and style images, it chooses the closest style image as the target for this specific panel. After that, the style transfer is applied to the corresponding masked images. Once all panels on the same page are transferred, the blending process reconstructs the style transferred images to get the final panel sequence. This reconstructed panel sequence is then mapped to a generated layout to get the resulting comic page. 

\subsubsection{Selection of study images}
The compositions of a comic panel have a lot of diversity because of the changes in the number of objects, character actions, view angles, and the placement of text boxes. Therefore, we took the camera shots concept in films as the basis of how to divide compositions. Unlike other style transfer problems, we consider the style of comics on a book scale rather than take only an image. For each comic book, we considered all the panels of the same style and then chose panels with 6 different compositions as representative images of the style. We took the combination of close, medium shots with various numbers of objects in the panel image.


\subsubsection{Content Similarity}
To address the bias of style transfer algorithms on texture features and focus on features of content, we use image hashing which to identify the similarities between image structures. We implement average hash \cite{zauner2010implementation} to down-sample the content- and style-masked images. This is then used to retrieve similar style images in terms of structure and image compositions to the content image. 

\subsubsection{Blending}
The resulting style transferred images from the masked channels are then combined to produce the final output. The blending is done in the order of the background image, followed by foreground image, and finally the text bubbles.

\subsubsection{Comic layouts}
Comic layouts include multiple panels organized on a page with different layout parameters. For reconstructing the output panel in terms of a target layout, we first consider the number of panels on the page. This information is then used to map to a layout in the target style that corresponds to the number of panels. More sophisticated panel layouts can be mapped if panel sizes can be adjusted. This becomes problematic due to artifacts created by stretching the original images to fit the new panel shapes. For simplicity in this first iteration, we choose layout mapping according to number of panels.

\section{Experiments and Evaluation}
Our experimental evaluation is structured as follows. First, we take single panels and entire pages, and directly apply currently available image style transfer framework to set as a baseline. Next, we create style-transferred images with variants that include a final blended page with all components and with ablated versions that exclude mask channels. This allows us to visually see the contribution of different channels in the transfer process. Finally, we run a visual story cloze test to evaluate the differences between coherence scores between a sequence of panels before and after the style transfer process. Table~\ref{common_art_style} shows an example of input and output images and layouts for each treatment.

\subsection{Single Panel Transfer}
Our first run of the algorithm focuses on creating a baseline with art styles that are highlighted in currently popular style transfer work. We choose the art style used in the starry image frequently used in other style transfer research as the target to ensure the quality of our implementation and to also highlight the challenges on our content. In the first three rows of table \ref{common_art_style}, we show the content image, target art style, single panel transfer and page transfer.  We notice in the starry night example (\emph{AS,N-M} in Table 1) that the texture stroke is preserved in the resulting images. As expected, it treats the content image as a whole and does not distinguish between channels. In the next example (\emph{AS,M} in Table 1), the model applies the dotted texture of the target style the color features from both foreground and background mix. The model treats text and text bubbles as a visual feature and also mixes its features in the visual scene.

Splitting the panels into foreground, background, and textbox. In the second experiment, we applied for the style transfer on two types of masks; one is the rectangle mask that we can parse from the dataset's annotations, the textboxs', objects' position; the other is the mask that fit the objects' shape. Figure  \ref{common_art_style} showed the results of using various masks. Also, in the last column, we exhibit the results of using layout to combine several transferred panels into a page. The reference layout was a comic page that had the same number of panels as the content page. For each row of the selected layout, the panel widths were adjusted to fit content manga panels, and the content panels will be centered to the target positions.


The results show that the foreground or background color distribution transfers well, as expected (\emph{CP-NM} in Table 2). Without masks there is noise due to mixing of features that affects the rendering of text bubbles. With the text mask, text is clear to read and arranged according to the style of comic text bubbles (\emph{CP-RM and CP-FM} in Table 2). For full pages, blending separately style transferred panels and adding them in the new layout yields better results. 

\begin{table*}
\centering
\begin{tabular}{|p{3.3cm}|p{3.3cm}|p{3.3cm}|p{3.3cm}|}
    \hline
      Content panel & Content page & Layout reference & Adjusted layout\\
    \hline 
      \includegraphics[width=\linewidth]{Images/5_5_panel.jpg}
      &\includegraphics[width=\linewidth]{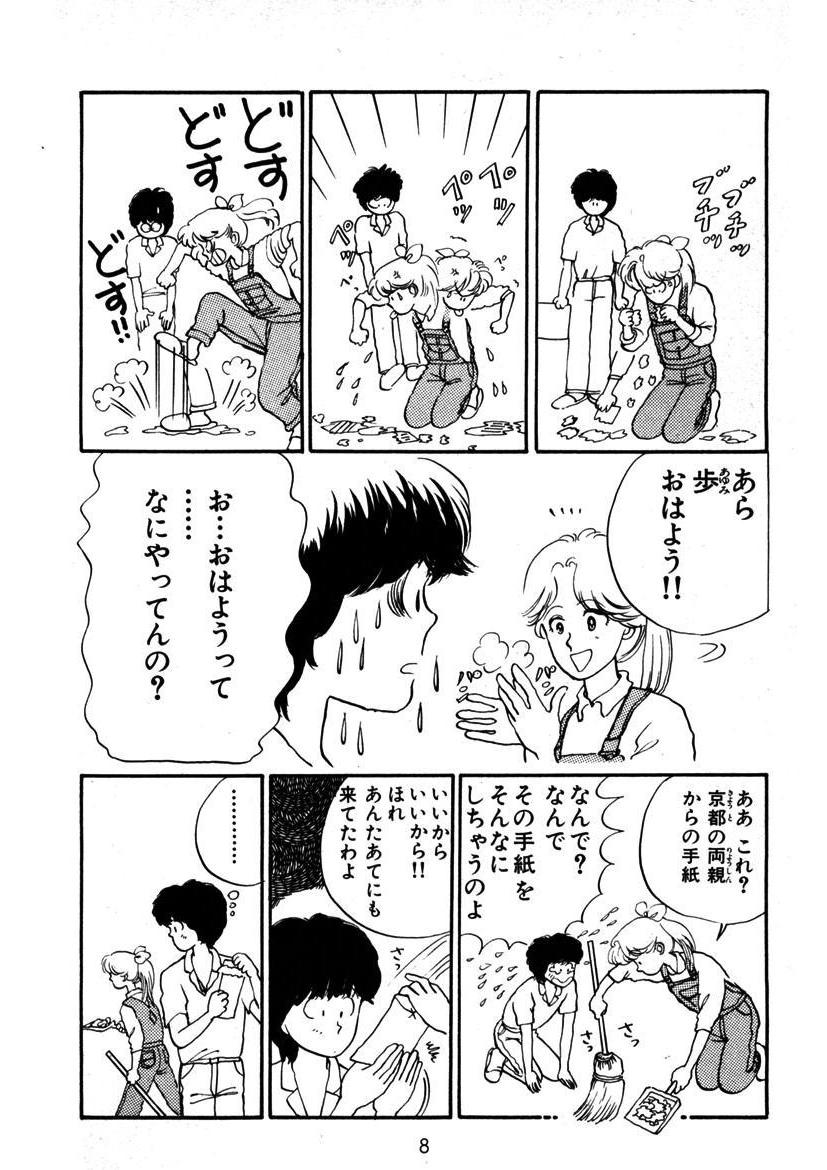}
      & \includegraphics[width=\linewidth]{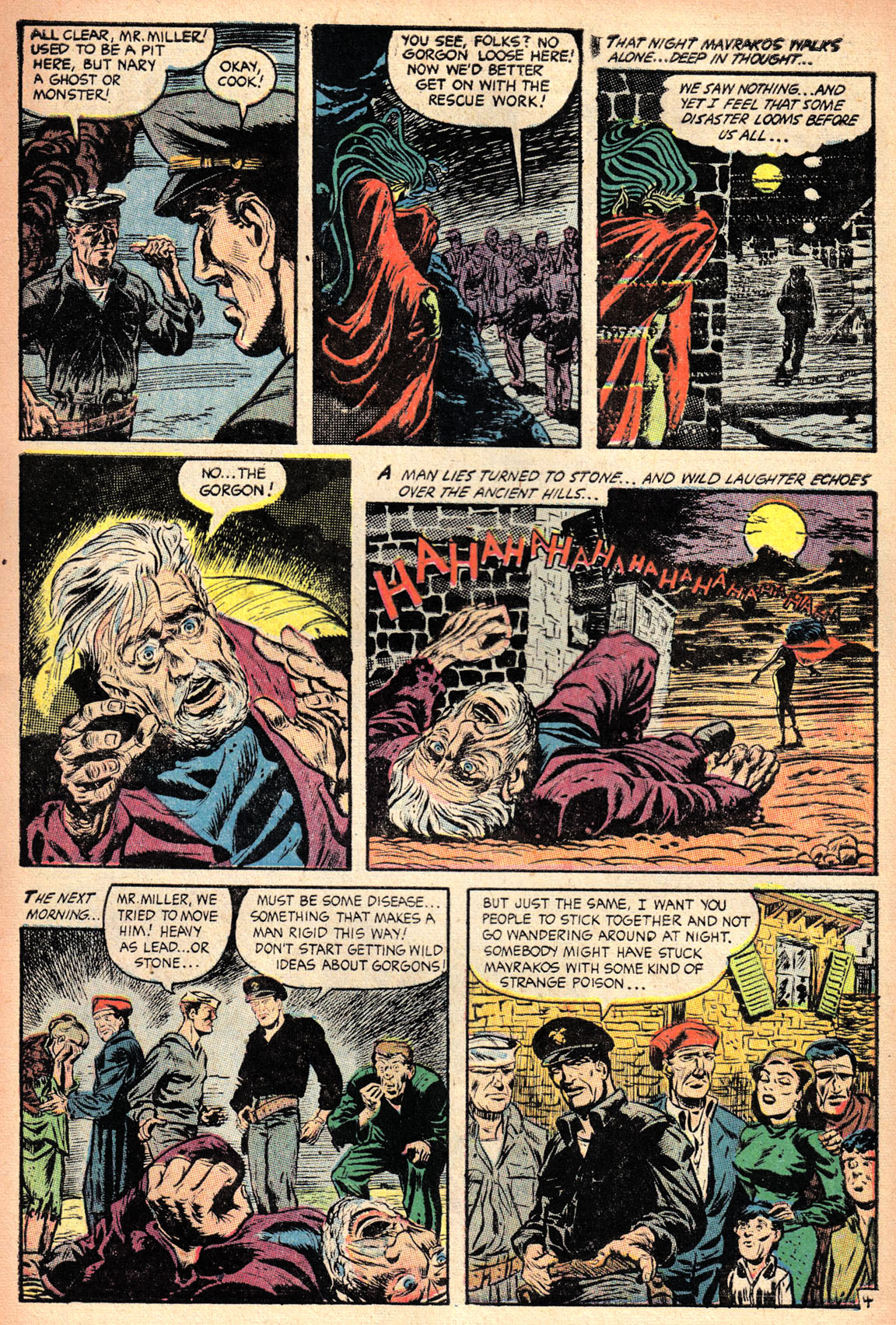} 
      &  \includegraphics[width=\linewidth]{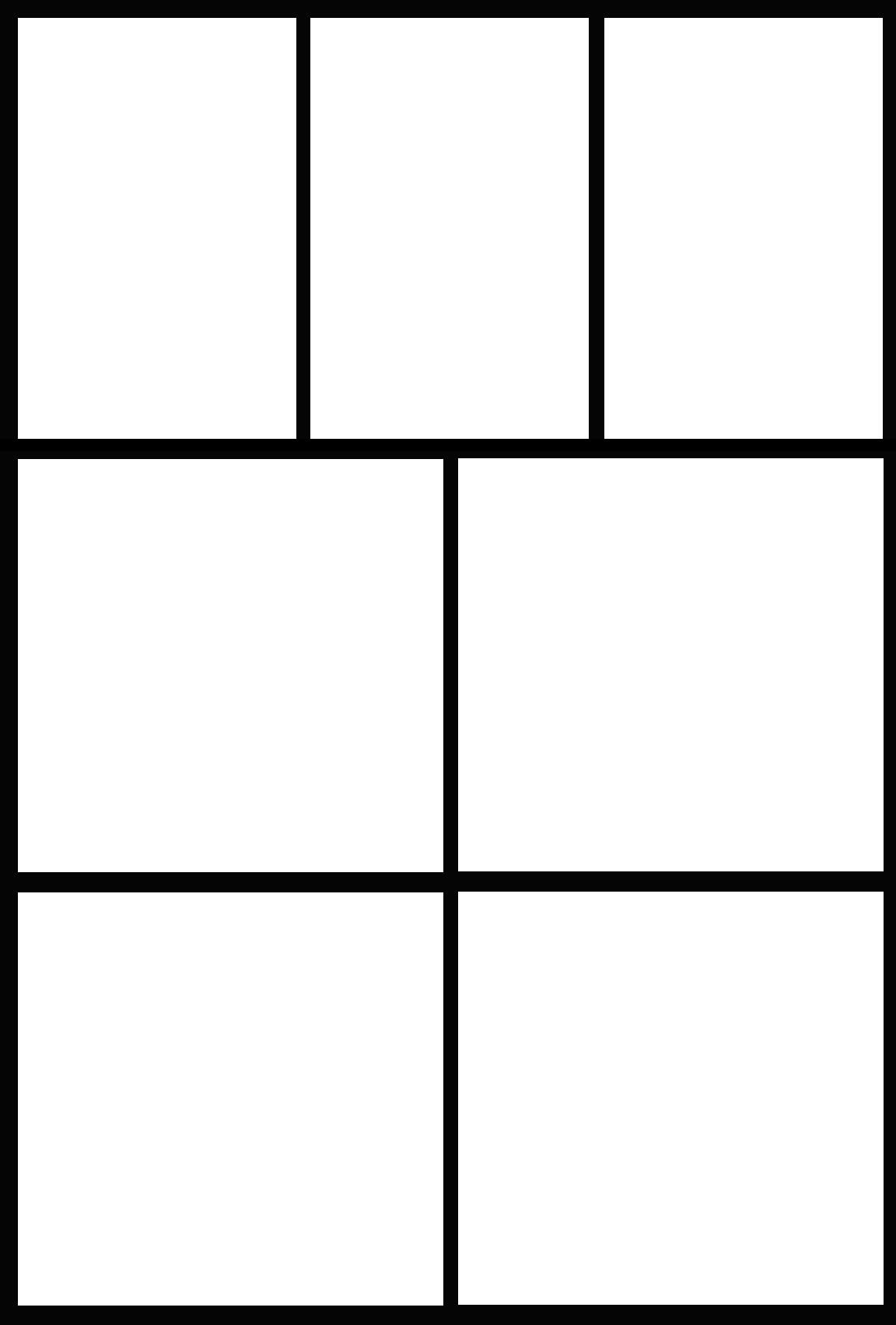} \\
    \hline 
      Settings & Style  & Panel & Page\\
    \hline     
   $AS$, $N\_M$& 
   \includegraphics[width=\linewidth]{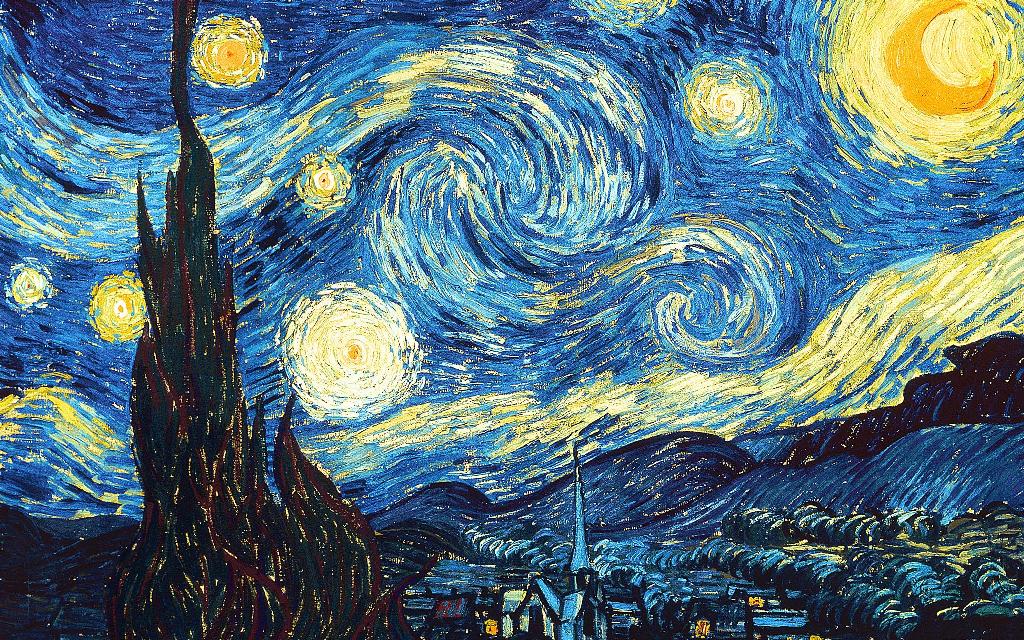} 
    &\includegraphics[height=\linewidth]{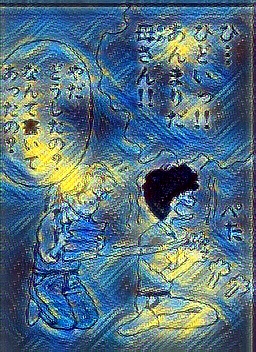}
   & \includegraphics[height=\linewidth]{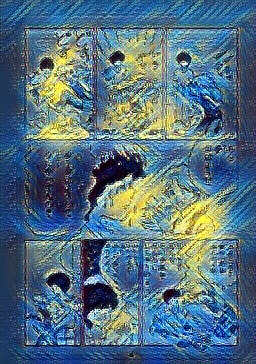}\\
    \hline     
    $AS$, $M$ 
    &
    \centering
    \includegraphics[height=0.4\linewidth]{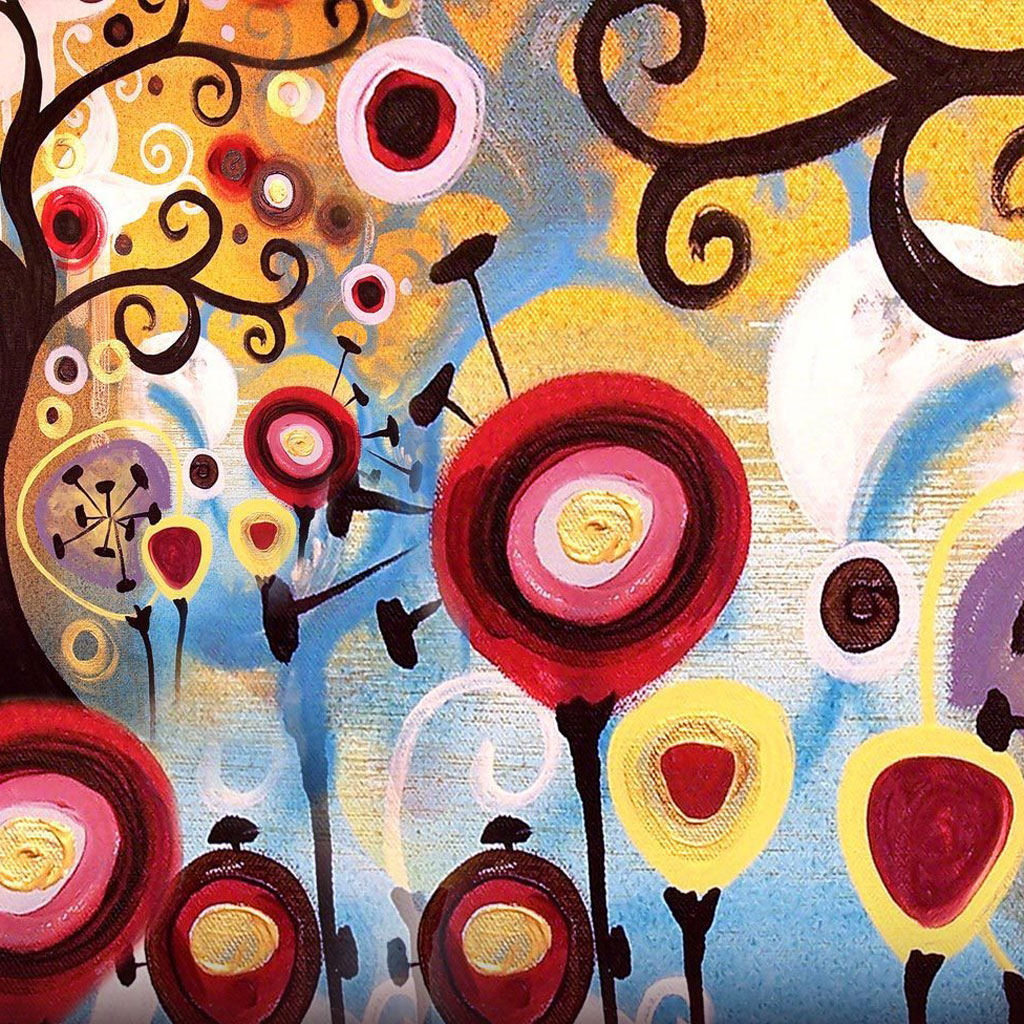} 
    \includegraphics[height=0.4\linewidth]{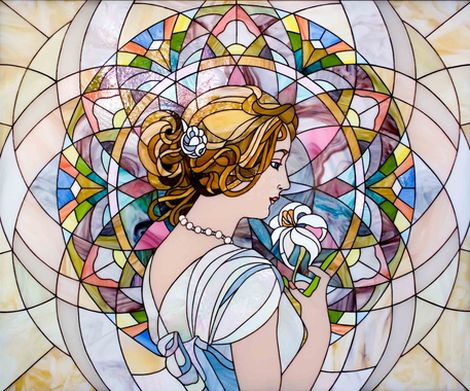}  
    &\includegraphics[height=\linewidth]{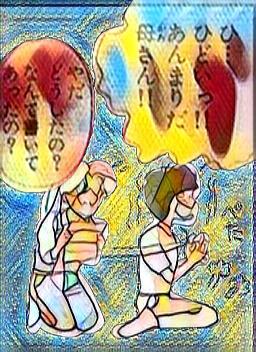}
    &\includegraphics[height=\linewidth]{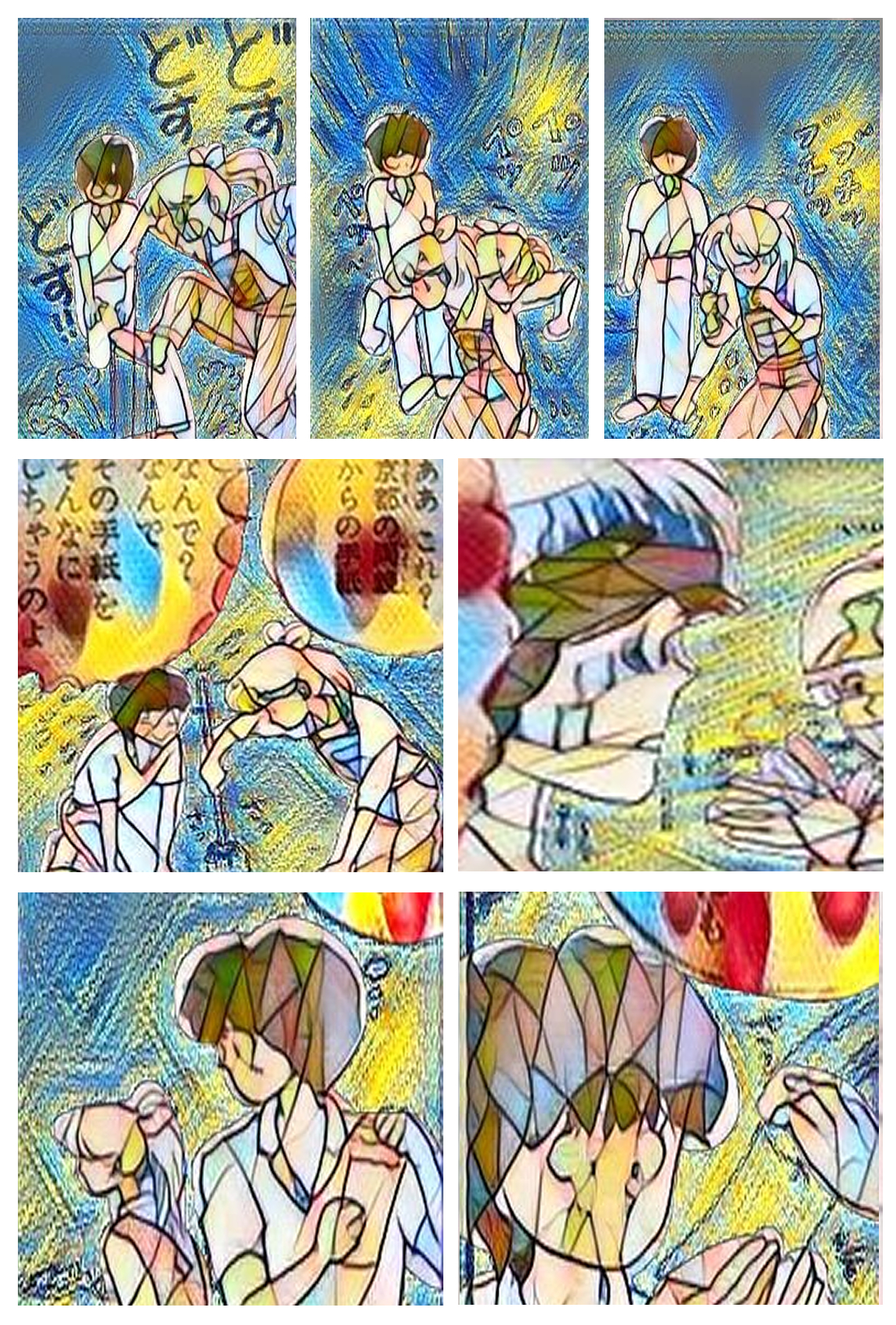}\\
    
    \hline 
    $CP$, $N\_M$ &\includegraphics[width=\linewidth]{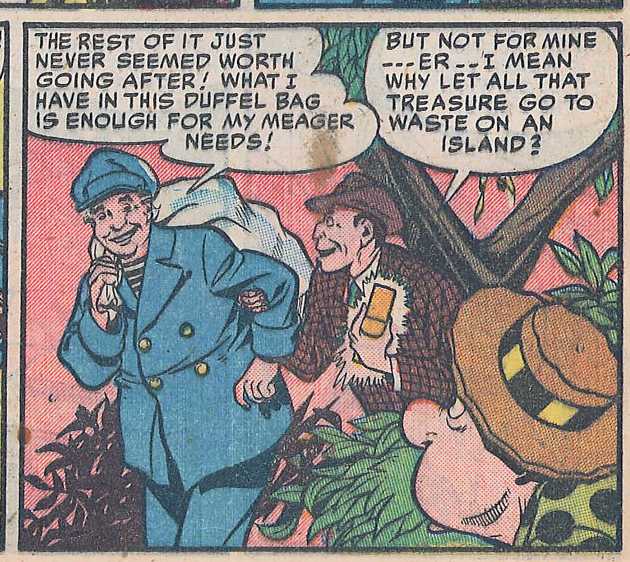} 
    &\includegraphics[height=\linewidth]{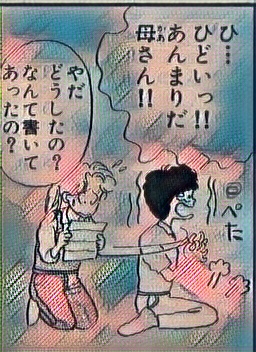}
    
    & \includegraphics[height=\linewidth]{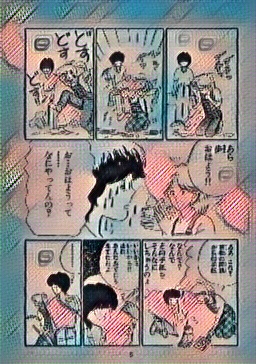}\\
    \hline 
    $CP$, $R\_M$ &\includegraphics[width=\linewidth]{Images/0_27_6.jpg} 
     & \includegraphics[height=\linewidth]{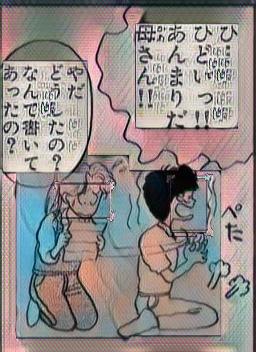}
     & \includegraphics[height=\linewidth]{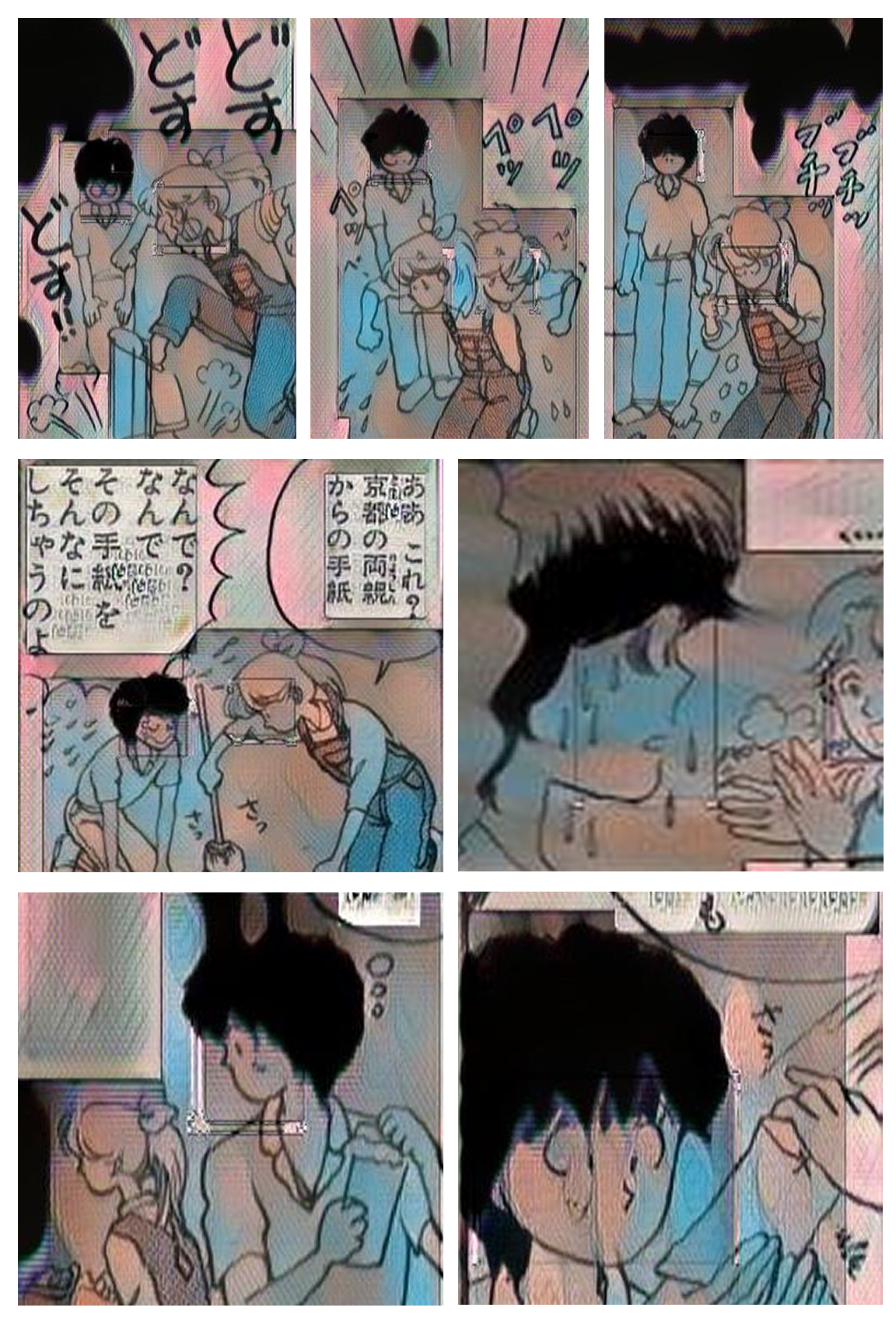}\\
    \hline     
    $CP$, $F\_M$ &\includegraphics[width=\linewidth]{Images/0_27_6.jpg} 
    &
    \includegraphics[height=\linewidth]{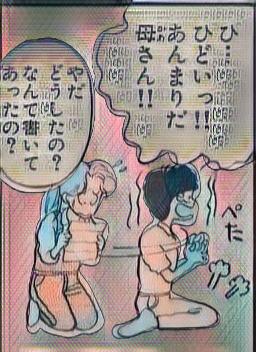}
    &\includegraphics[height=\linewidth]{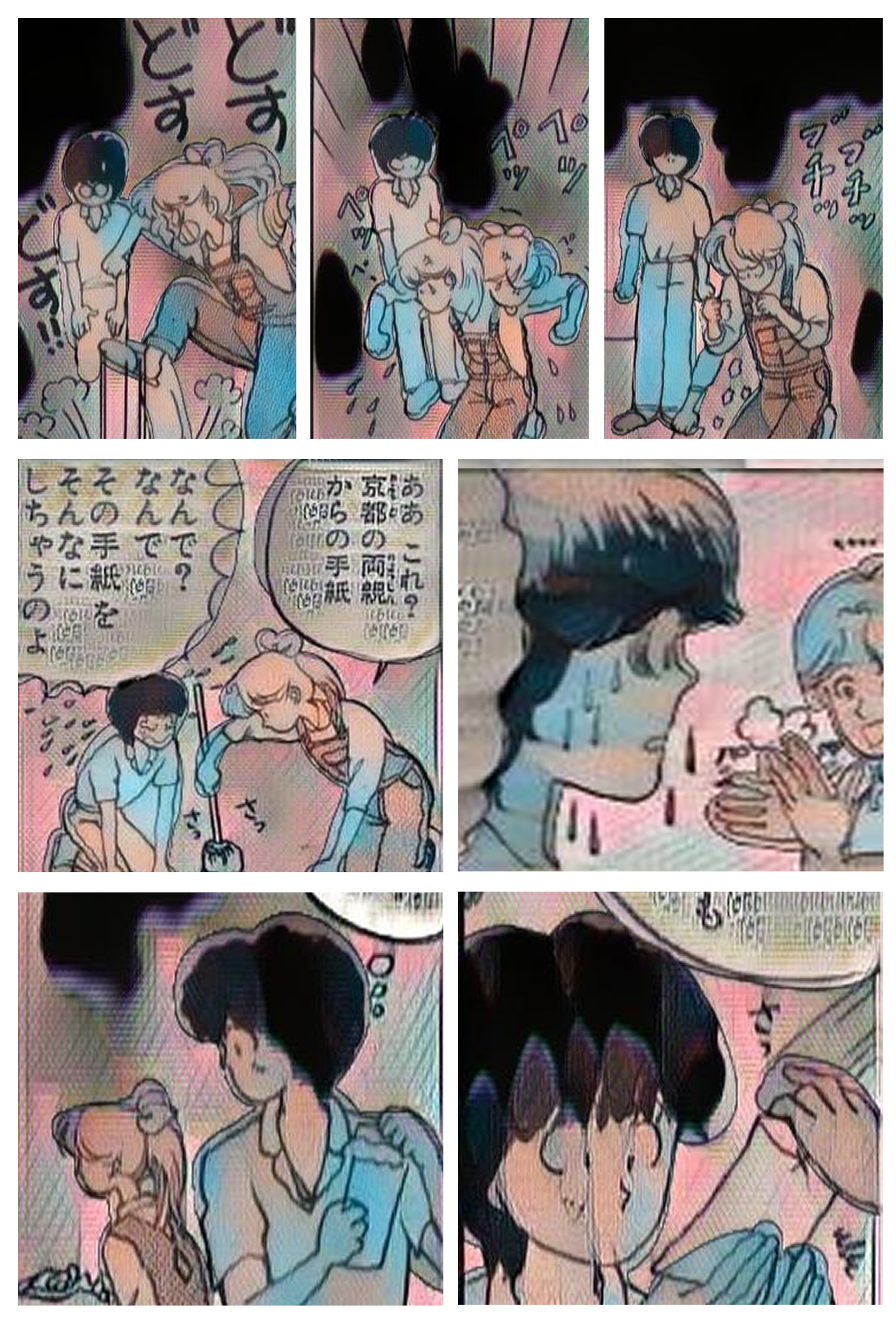}\\
    \hline     
    \hline     
\end{tabular}
    \caption{
    \label{common_art_style} \small{The comparison between using art paintings as target style and using a content-rich image as target style. The settings are: with art style but no masks (AS, $N\_M$), with art style and masks (AS, $M$), with comic panel but no masks (CP, $N\_M$), with comic panels and rectangle masks (CP, $R\_M$), with comic panels and fit masks (CP, $F\_M$).}
    }
\end{table*}

 

\subsection{Panel Sequence and Layout}
Our next experiment focuses on the coherence of sequences in terms of content. Our experiment is designed test whether coherence is influenced by style transfer given a sequence of consecutive panels. The following subsections described how we design our comparison.

In this experiment, we transferred the style for 8 volumes of manga, 29,039 panels in total. These form the content of the visual story cloze task. 

\begin{table}
\centering
\begin{tabular}{|p{3cm}|p{2cm}|p{2cm}|}
    \hline
     Settings & feature-C & feature-M \\
    \hline 
    N-T & 48.2 & 70.6\\
    \hline     
    T-W & 50.4 & 65.2 \\
    \hline     
    T-M & 51.7 & 67.1\\
    \hline     
    T-C & 53.3 & 69.0 \\
    \hline     

\end{tabular}
    \caption{
    \label{tab:ClozeR} \small{The cloze-test accuracy (the rate that our model got correct answers out of all questions) comparison. This showed two experiment groups, and each contained four sets. The groups used different fine-tuned features: one trained on COMICS, the other trained on Manga109. The 4 settings are no style transfer (N-S), style transfer with the whole image (T-W), style transfer with masks (T-M), and style transfer with composition features (T-C) }
    }
\end{table}

\subsubsection{Cloze task}
Closure is a process that involves understanding of individual panels and making connective and often complex inferences between consecutive panels. To characterize the differences in the performance of the computational model developed for COMICS on the Manga109 dataset, we employed two cloze tasks: text cloze and visual cloze~\cite{iyyer2017amazing}. A model’s ability to correctly predict the next panel in a given a sequence of panels of context is tested through these tasks. In our case, we use the cloze test on original sequence of images and compare the output with the style transferred image sequence. Given a sequence of panels as context, the model is asked to predict the most likely ending out of 3 candidates. While a comic panel is a combination of image and text content, we employed the visual tasks only. In this case, the model only observes visual features for prediction. Text features would be relevant but require proper translation or Japanese language model. This avenue can be explored in future work. 

\subsubsection{LSTM model}
To answer the visual-cloze tasks, we employed an LSTM based comprehension model. It encodes the context images with the 4096-d fc7 layer of VGG-16. The fc7 features of each panel are feed to an LSTM. The model converts the sequence of context panels into fixed-length vectors and scores the answer candidates by taking the inner product of the candidates with the vectors then normalizing it with the softmax function. The model projects both of the answer candidates and the context to 512-d representation. The feature descriptor we used in this experiment was a VGG16 pre-trained on image net data and then fine-tuned with either COMICS  or Manga109. Therefore, our experiment has two sets: one used the features fine-tuned with COMICS to describe manga images, the other used the features with Manga109. 

\subsubsection{Results}
We divide the cloze-test into two groups, and each group had four sets. The groups used different feature descriptors to encode image context. The four sets are split according to whether they have style transfer, whether they used masks, and whether the target style chosen according to image composition? 

Our target-style was comic panels from COMICS that represent Western comics, and the content images all came from Japanese manga books. The first row of Table \ref{tab:ClozeR} shows that the feature tuned on COMICS didn't capture the manga image as well as the features tuned on Manga109. The former's accuracy on the cloze-test is much lower than the latter's when the content images had no style transfer.

The results in both groups suggested that the transferred image still preserved the narrative characteristic to some extent. Because with the new data, the comprehension model can still predict the right answers for visual-cloze tasks. Besides, the two groups also show some interesting phenomena.

The results in the first column suggest a trend that when the manga images preserve more content after style transfer (to western comic style), the accuracy increases accordingly. When the whole panel image transferred to comic style, the accuracy slightly increased. After the masks based on foreground, background, and so on were added to images, the accuracy increases slightly again. And then, after the model chose style according to compare the similarity between image composition, the accuracy increase once again.

In the study with manga features (second column), there is a drop in accuracy with whole image style transfer but improvement with the application of masks and composition features.

\section{Conclusions and Future Work}

We presented the research problem of style transfer between comics and manga that represent multi-modal sequential narrative media. We described the feature set for this medium, highlighted challenges in terms of the interaction of layout, text, and scene features, and proposed a transfer framework. We introduced content masking with parallel style transfer for independent features and illustrated the differences between various single-image compositions and combined the panels through an adjusted layout on both individual panels and full pages. To address the narrative communication aspect, we propose comprehension preserving transfer and evaluate style transfer modules based on visual cloze tests for narrative understanding. Results shown in this paper highlight the challenges for this medium and also set a baseline for future work on this topic. 

As a new perspective on style transfer algorithms, we hope to engage the narrative research community on interesting discussions around richer aspects of style transfer. This work also potentially has impact on future work related to interactivity in terms of visual interactive games and player styles and preferences as well as dynamic panel transitions. Our initial motivation for comic analysis was to study the use of composition for dynamic camera control in narrative-based environments.

\bibliographystyle{natdin}
\bibliography{reference}

\end{document}